\documentclass[10pt,twocolumn,letterpaper]{article}

\usepackage{algorithm}
\usepackage{algorithmic}
\usepackage{multirow}
\usepackage{arydshln}
\usepackage{booktabs}
\usepackage{diagbox}  %
\usepackage{afterpage} %
\usepackage{fontawesome5} %
\usepackage{pifont}
\usepackage{svg} 
\usepackage{tikz}

\usepackage{sourcesanspro}

\newcommand{\modelname}[1]{\textcolor{nice-model}{\underline{{Subjective Camera 1.0}}}}

\usepackage{iccv}              %

\usepackage{xcolor}
\colorlet{dark-blue}{blue!50!black}
\colorlet{dark-cyan}{cyan!75!black}
\colorlet{dark-purple}{purple!50!black}
\colorlet{dark-red}{red!75!black}
\colorlet{dark-green}{green!80!black}
\colorlet{dark-orange}{orange!50!black}
\colorlet{dark-gray}{black!75}
\colorlet{light-gray}{black!30}
\definecolor{nice-red}{HTML}{E41A1C}
\definecolor{nice-orange}{HTML}{FF7F00}
\definecolor{nice-yellow}{HTML}{FFC020}
\definecolor{nice-green}{HTML}{39b54a}
\definecolor{nice-blue}{HTML}{0071bc}
\definecolor{nice-purple}{HTML}{984EA3}

\definecolor{nice-model}{HTML}{6600CC}

\definecolor{darkgreen}{HTML}{3F7D31}
\definecolor{darkred}{HTML}{BA3132}

\definecolor{second}{rgb}{1, 0.85, 0.7}
\definecolor{best}{rgb}{1, 0.7, 0.7}
\definecolor{third}{rgb}{1,1, 0.8}

\colorlet{verylight-gray}{black!10}
\definecolor{LightCyan}{rgb}{0.66,0.85,0.76}

\newif\ifsymbol
\symbolfalse

\ifsymbol
\newcommand{\whu}{{\ding{95}}}%
\newcommand{\hubei}{\ding{168}}%
\newcommand{\zhongguancun}{\ding{169}}%
\newcommand{\ailab}{\faFlask}%
\else
\newcommand{\whu}{1}%
\newcommand{\hubei}{2}%
\newcommand{\zhongguancun}{3}%
\newcommand{\ailab}{4}%
\fi

\newcommand{\shuttericon}{\includegraphics[width=0.8em]{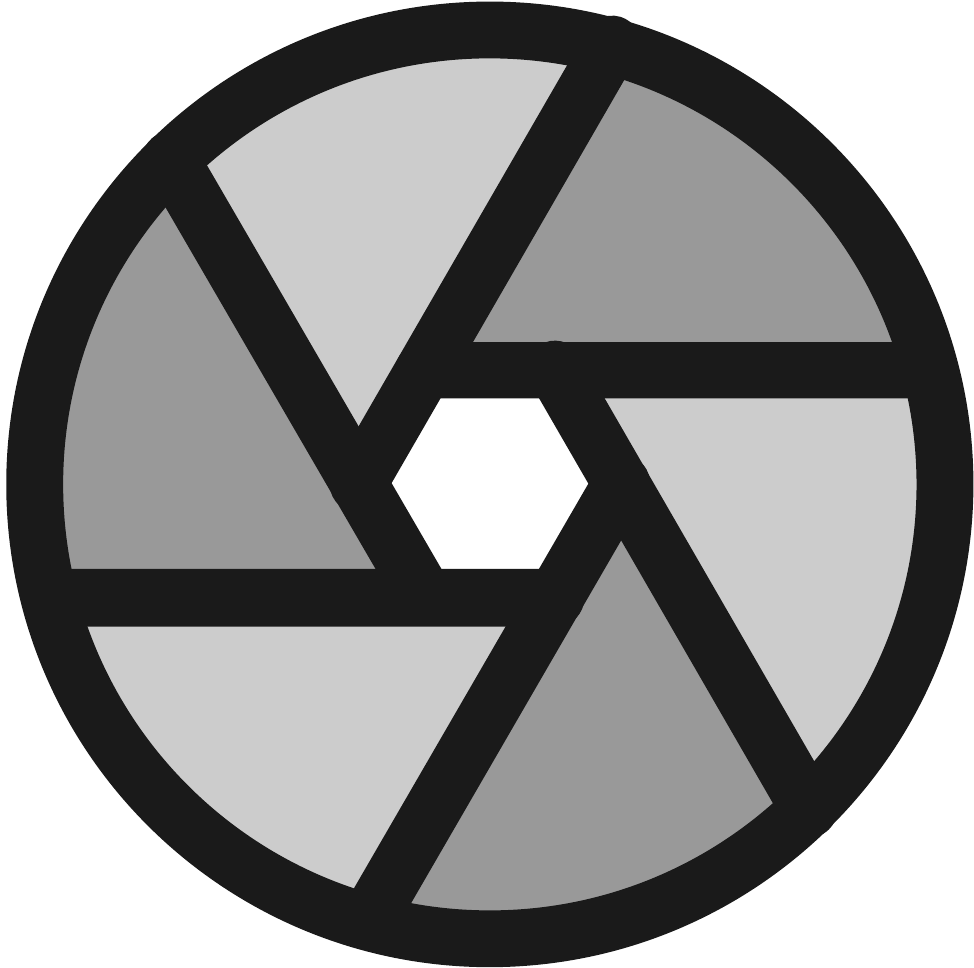}}

\usepackage{colortbl}  
\definecolor{Green}{cmyk}{0.059, 0, 0, 0}
\definecolor{Gray}{cmyk}{0.06, 0, 0.151, 0.012}

\definecolor{sketch1}{HTML}{8C44F4}
\definecolor{sketch2}{HTML}{C0FDA4}
\definecolor{sketch3}{HTML}{00B0F0}
\definecolor{sketch4}{HTML}{F2AA84}

\newcommand{\colcirc}[3][0.8ex]{%
    \raisebox{0.5pt}{%
        \tikz{\node[draw=#3,fill=#2,circle,scale=1,minimum size=#1,inner sep=0pt](){};}%
    }%
}
\robustify{\colcirc}

\newcommand{\heading}[1]
{
\vspace{1mm}\noindent\textbf{#1}
}

\definecolor{iccvblue}{rgb}{0.21,0.49,0.74}
\usepackage[pagebackref,breaklinks,colorlinks,citecolor=dark-blue, linkcolor=dark-red]{hyperref}

\def\paperID{10998} %
\def\confName{ICCV}
\def\confYear{2025}

\title{\includegraphics[width=0.8cm]{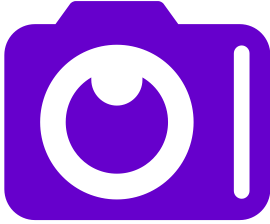}~\textcolor{dark-gray}{\modelname{}}: Bridging Human Cognition and Visual Reconstruction through Sequence-Aware Sketch-Guided Diffusion}

\makeatletter
\def\@fnsymbol#1{\ensuremath{\ifcase#1\or \text{\faUncharted} \or \text{\faEnvelope[regular]} \or \ddagger\or
		\mathsection\or \mathparagraph\or \|\or **\or \dagger\dagger
		\or \ddagger\ddagger \else\@ctrerr\fi}}
\makeatother

\ifsymbol
\author{
\textsuperscript{\whu,\hubei,\zhongguancun}Haoyang Chen\thanks{These authors contributed equally to this work.} \quad
\textsuperscript{\whu,\hubei}Dongfang Sun\footnotemark[1] \quad
\textsuperscript{\whu,\hubei}Caoyuan Ma\footnotemark[1] \quad
\textsuperscript{\whu,\hubei}Shiqin Wang
\textsuperscript{\whu,\hubei}Kewei Zhang \\
\textsuperscript{\whu,\hubei,\zhongguancun}Zheng Wang\thanks{Corresponding authors.} \quad
\textsuperscript{\ailab}Zhixiang Wang\footnotemark[2] \\
\normalsize{\textsuperscript{\whu}National Engineering Research Center for Multimedia Software, Institute of Artificial Intelligence, School of}\\
\normalsize{Computer Science, Wuhan University \quad \textsuperscript{\hubei}Hubei Key Laboratory of Multimedia and Network Communication Engineering}\\
\normalsize{\textsuperscript{\zhongguancun}Zhongguancun Academy, Beijing, China. 100094 \quad \textsuperscript{\ailab}{CyberAgent AI Lab, Japan} }
}
\else
\author{
Haoyang Chen\textsuperscript{\whu,\hubei,\zhongguancun}\thanks{These authors contributed equally to this work.} \quad
Dongfang Sun\textsuperscript{\whu,\hubei}\footnotemark[1] \quad
Caoyuan Ma\textsuperscript{\whu,\hubei}\footnotemark[1] \quad
Shiqin Wang\textsuperscript{\whu,\hubei} \quad
Kewei Zhang\textsuperscript{\whu,\hubei} \\
Zheng Wang\textsuperscript{\whu,\hubei,\zhongguancun}\thanks{Corresponding authors.} \quad\quad
Zhixiang Wang\textsuperscript{\ailab}\footnotemark[2] \\
\normalsize{\textsuperscript{\whu}National Engineering Research Center for Multimedia Software, Institute of Artificial Intelligence, School of}\\
\normalsize{Computer Science, Wuhan University \quad \textsuperscript{\hubei}Hubei Key Laboratory of Multimedia and Network Communication Engineering}\\
\normalsize{\textsuperscript{\zhongguancun}Zhongguancun Academy, Beijing, China. 100094 \quad \textsuperscript{\ailab}{CyberAgent AI Lab, Japan} }
}
\fi

\begin{document}

\thispagestyle{empty}
\twocolumn[{%
\renewcommand\twocolumn[1][]{#1}%
\tabcolsep=5pt
\maketitle
\setcaptiontype{figure}
\centering
\includegraphics[width=0.95\textwidth]{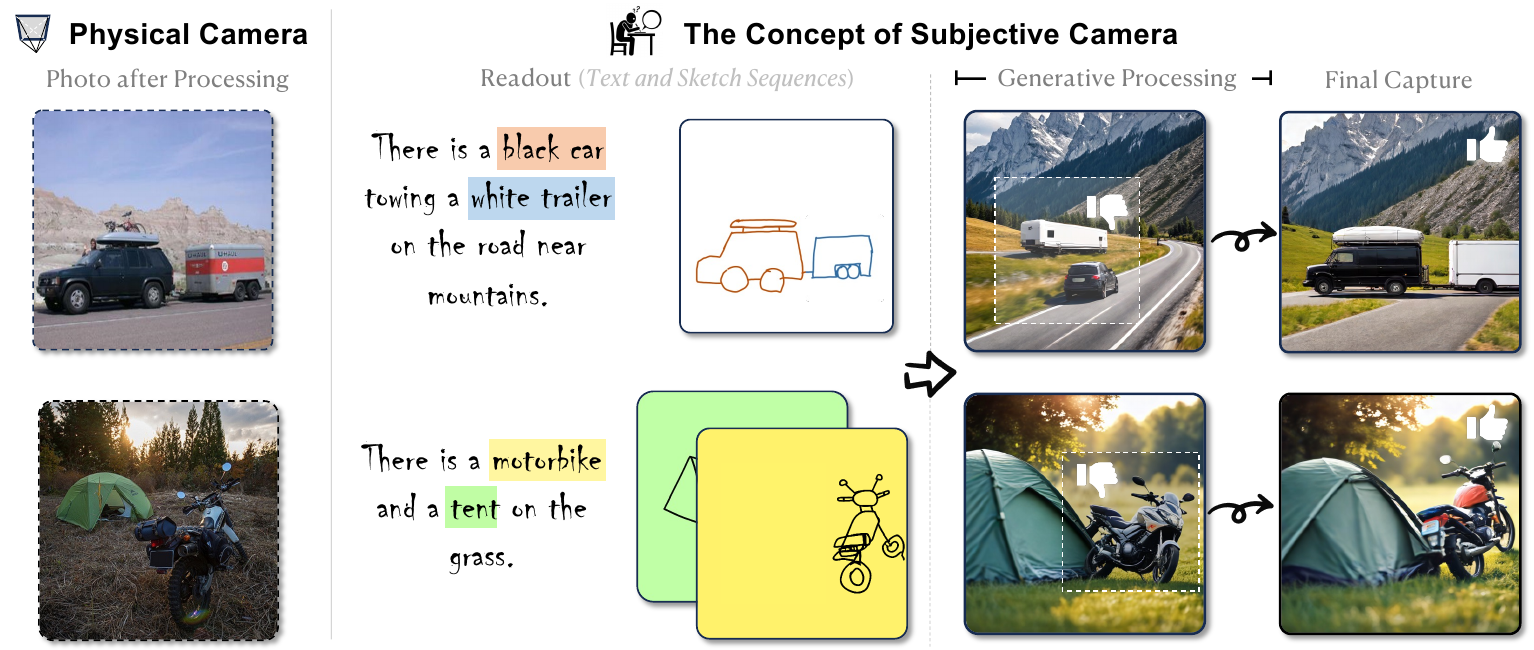}
\setcounter{figure}{0}
\captionof{figure}{
\textbf{What is a subjective camera?}
A subjective camera transforms a person’s memory into photorealistic images. We take a small yet crucial step toward this vision by leveraging generative models to sequentially decode textual and sketch‑based readouts in their natural order, paving the way for reconstructing moments without physical cameras.
}
\label{fig:intro_motivation}
\vspace{10mm}
}]

\renewcommand{\thefootnote}{\fnsymbol{footnote}}  %
\setcounter{footnote}{0}                          %
\footnotetext[1]{These authors contributed equally to this work.}
\footnotetext[2]{Corresponding authors.}

\renewcommand{\thefootnote}{\arabic{footnote}}

\begin{abstract}

We introduce the concept of a \emph{subjective camera} to reconstruct meaningful moments that physical cameras fail to capture. We propose \emph{Subjective Camera 1.0}, a framework for reconstructing real‑world scenes from readily accessible subjective readouts, \ie, textual descriptions and progressively drawn rough sketches. Built on optimization‑based alignment of diffusion models, our approach avoids large‑scale paired training data and mitigates generalization issues. To address the challenge of integrating multiple abstract concepts in real‑world scenarios, we design a Sequence‑Aware Sketch‑Guided Diffusion framework with three loss terms for concept‑wise sequential optimization, following the natural order of subjective readouts.
Experiments on two datasets demonstrate that our method achieves state‑of‑the‑art performance in image quality as well as spatial and semantic alignment with target scenes. User studies with 40 participants further confirm that our approach is consistently preferred.
Our project page is at: \href{subjective-camera.github.io}{subjective-camera.github.io}

\end{abstract}
    
\section{Introduction}
\label{sec:intro}

Cameras have long been the most popular devices for preserving memorable moments.
However, a physical camera cannot always be present to capture \emph{every} moment. This raises a question of how such unrecorded experiences can be preserved as \emph{pixels} in the absence of a camera.

This paper proposes the concept of a {\emph{subjective camera}}\footnote{While the term ``subjective camera'' appears in cinematic contexts, our usage here differs significantly in meaning.}, where humans function as ``imaging devices'', aiming to faithfully reconstruct real-world scenes from memory.
{Specifically, we define a subjective camera as a system through which individuals encode sensory inputs into memory based on personal salience and emotional context, and later read out these stored representations and decode them into pixels that reconstruct scenes they experienced. We draw an analogy to physical cameras: just as a camera records, reads out, and reconstructs the visual world through optical, electronic, and computational processes, the human mind selectively captures (``records'') perceptual details, which are organized into a retrievable format (``read out'') and later reconstructs them as pixels.

It is generally challenging for humans \emph{without} professional training to directly reconstruct a dense, pixel‑wise image. Instead, the most practical approach is to \emph{read out} the mental scene through textual descriptions and/or freehand sketches and leverage computational tools to decode these readouts. However, text alone often fails to fully convey a mental image, particularly in terms of object layout and fine details. Freehand sketches provide complementary information beyond text but remain sparse and often contain noise and uncertainty~\cite{HuSigir, lin2023beyond}. With the advancement of generative models~\cite{rombach2022high,peebles2023scalable}, this long‑standing vision is becoming increasingly feasible by decoding such subjective readouts into photorealistic imagery.

Although not intended to reconstruct real-world scenes, several works have explored generating images from combined text and sketch inputs~\cite{wu2023sketchscene, zhangTy2024sketch,zhao2023uni,cheng2024scene,zhang2024sketchc,sarukkai2024block}.
These approaches either adapt model weights using \emph{large-scale} paired data~\cite{voynov2023sketch,zhang2023adding, mou2024t2i, wu2023sketchscene, zhangTy2024sketch,zhao2023uni,sarukkai2024block} or modify the latent variables through \emph{per-scene} optimization~\cite{voynov2023sketch,cheng2024scene,zhang2024sketchc}.
Training‑based methods directly learn the mapping from sketches to pixels. However, they require extensive paired data and struggle with user‑specific biases, often generalizing poorly to abstract sketches beyond the training distribution. In contrast, training‑free methods avoid the heavy data requirements, high computational costs, and generalization challenges of training‑based approaches.

However, optimization-based methods often fail to reconstruct \emph{multiple} concepts from \emph{idiosyncratic} sketches, as would be required for recreating real‑world scenes.
This often leads to missing concepts or misaligned spatial relationships (see Figure~\ref{fig:motivation}\textcolor{dark-red}{a}).
The main obstacle lies in simultaneously interpreting multi-concept sketches,  which inherently combine \emph{varying} levels of abstraction and randomness across different concepts. For instance, freehand sketches may only roughly indicate object boundaries, omit fine details such as texture and geometry, or distort proportions in non‑standard ways.
While generative priors can fill sparse regions and mitigate randomness during the transformation from sketches to pixels, attempting to integrate all concepts at once often reduces the problem to a least‑squares solution~\cite{bar2023multidiffusion,cheng2024scene}, ultimately blurring their distinct characteristics.

Our key idea is simple --- to apply optimizations to each concept \emph{individually} and do so \emph{sequentially}.
This design considers not only 
the limitations of the ``snapshot'' generation process but also human cognition. Humans tend to describe their mental images step‑by‑step \cite{kosslyn1988aspects}, typically starting with a textual description of abstract concepts or the overall scene structure, and then progressively refining individual elements with increasing detail (see Figure~\ref{fig:motivation}\textcolor{dark-red}{b}).

\begin{figure}[t]
    \centering
    \includegraphics[width=0.8\linewidth]{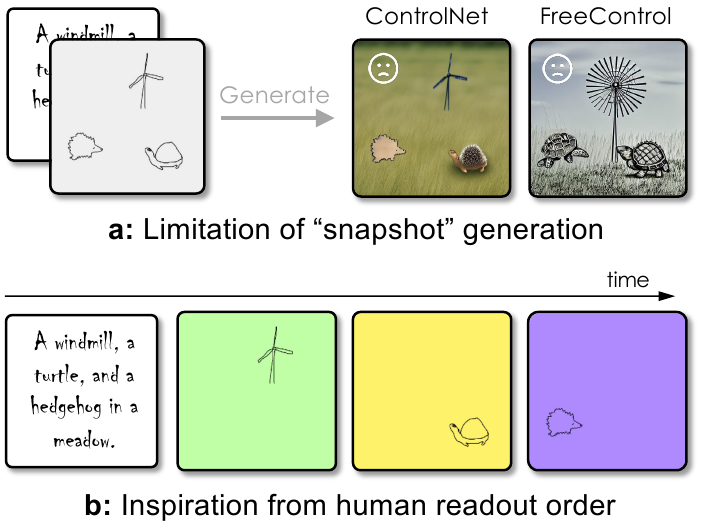}
    \caption{\textbf{Motivation for concept‑wise sequential optimization.} 
    (a) Attempting to ``snap'' all concepts into a complete image at once using generative diffusion priors often leads to suboptimal results, such as distortions and misaligned semantics or layouts. 
    (b) In contrast, humans reconstruct mental scenes step by step. This natural readout order inspires our design\protect\footnotemark, making it better suited to leverage current generative diffusion priors.
    }
    \label{fig:motivation}
\end{figure}
\footnotetext{This design philosophy is reminiscent of the progressive decoding used in rolling‑shutter image sensors, which arises from the need to accommodate the limited processing capacity of the circuits.}

Building upon this idea, we introduce a sequence‑aware, sketch‑guided diffusion framework that sequentially optimizes the latent noise of a pre‑trained text‑to‑image (T2I) diffusion model~\cite{rombach2022high} to align individual subjective readouts.
The order of these optimizations mirrors the natural process of reading out subjective impressions: it begins with a text‑reward optimization (Section~\ref{sec:TRO}) to align the generated image with the initial textual description, followed by concept-wise sequential optimizations (Section~\ref{sec:OG}) to incorporate progressively provided sketches that may encode shapes, poses, and spatial arrangements. The sequential optimization also incorporates a loss to ensure that previously optimized concepts are not significantly altered.

Two additional components are proposed to improve the concept-wise sequential optimization.
First, each input sketch is individually encoded into the latent space to provide spatial guidance. To bridge the gap between rough sketch shapes and their corresponding real‑world forms, we introduce an optimization‑based inversion strategy for refining this spatial guidance (Section~\ref{sec:LO}).
Second, because input sketches often lack appearance details, directly optimizing with their guidance over multiple iterations may lead to unnatural textures and degraded visual fidelity. We therefore leverage the latent representation obtained from text‑reward optimization to define an appearance loss (Section~\ref{sec:SE}), thereby preserving visual style quality.

By processing sketches individually and sequentially, our approach naturally aligns with the human cognition process, enabling users to incrementally construct mental images while ensuring each addition respects and complements previous elements. This sequence‑aware generation strategy provides \emph{fine‑grained} control over complex multi‑concept scenes while preserving the user’s subjective intent encoded in the drawing order. Evaluations on two datasets demonstrate that our method achieves state‑of‑the‑art performance in recreating a real-world scene from human memory, demonstrating the potential of subjective cameras.
Notably, our framework is entirely \emph{training‑free}, eliminating the need for large‑scale paired datasets and avoiding generalization.

To sum up, our contributions are twofold:
\begin{itemize}[label=\shuttericon, leftmargin=0.65cm, topsep=0cm, partopsep=0pt, parsep=0pt, itemsep=0pt]
\item We propose the concept of a {subjective camera}, which functions the human as an ``imaging device'' and leverages computational tools to transform cognitive impressions into photographs.
\item We propose a sequence-aware diffusion-guided generation framework that enables faithful reconstruction of complex, multi‑concept scenes. This framework aligns with human cognition and surpasses ``snapshot'' generation methods, thereby establishing a new paradigm for cognition‑driven image generation.
\end{itemize}

\section{Related Work}
\label{sec:relatedWork}

\heading{Aligning Image Synthesis Models}
Despite their success, T2I generative models~\cite{rombach2022high,ramesh2021zero,peebles2023scalable} often fail to reproduce the fine‑grained semantics and compositional details described in complex prompts. Reward‑based alignment has emerged as a promising direction~\cite{kirstain2023pick,xu2023imagereward,sundaram2024cocono,eyring2025reno,guo2024initno,lee2024parrot,xie2025dymo}, using human preference models~\cite{xu2023imagereward,wu2023human,kirstain2023pick} to guide generation.
The human preference models were trained on paired human preference data. 
Early works~\cite{kirstain2023pick,xu2023imagereward} fine‑tune diffusion models with the reward models, whereas later methods~\cite{eyring2025reno,lee2024parrot} sidestep costly fine‑tuning by directly optimizing the latent noise~\cite{eyring2025reno,xie2025dymo}. However, these techniques largely improve global prompt adherence while neglecting \emph{spatial} control.
Our model not only aligns the T2I generative models with the text prompt, but also ensures the spatial information to respect to user-provided sketches.

\heading{Controllable Image Synthesis}
T2I models inherently rely on highly \emph{compressed} textual tokens, which limits their controllability and often prevents them from meeting user expectations~\cite{rombach2022high,ramesh2021zero,peebles2023scalable}. To overcome these limitations, a growing body of research has explored controllable generation by incorporating fine‑grained conditioning signals, such as reference images~\cite{brooks2023instructpix2pix}, edge maps or contours or skeletons~\cite{zhang2023adding}, semantic layouts~\cite{bar2023multidiffusion,zheng2023layoutdiffusion,chen2024training}, and lighting specifications~\cite{zhang2025scaling}. 
One line of work fine‑tunes pretrained diffusion models by adding trainable modules~\cite{zhang2023adding,zhao2023uni,mou2024t2i}, while others directly retrain the diffusion backbone by minimizing reconstruction objectives~\cite{brooks2023instructpix2pix}.
In parallel, training‑free techniques~\cite{sundaram2024cocono,guo2024initno,hertz2022prompt,epstein2023diffusion} have been developed to enhance controllability without modifying model weights. These include {attention injection} or {latent optimization} strategies, such as Prompt‑to‑Prompt editing~\cite{hertz2022prompt} and diffusion self‑guidance~\cite{epstein2023diffusion}. 
Despite these advances, existing methods often fail in complex scenes involving multiple guidance given by freehand sketches, where interactions between elements lead to interference and degraded synthesis quality. Building on these insights, our method incorporates attention‑based guidance extraction but \emph{extends} it to better handle multi‑concept freehand sketches.

\heading{Sketch-to-Image Synthesis}  
It is the task most relevant to our concept of the subjective camera. It aims to generate images from freehand sketches along with textual prompts.
Several recent efforts have attempted to adapt T2I diffusion models for sketch-to-image synthesis~\cite{voynov2023sketch,zhangTy2024sketch,cheng2024scene,zhang2024sketchc,mo2024freecontrol,sarukkai2024block}.
Training‑based approaches~\cite{zhangTy2024sketch,sarukkai2024block} fine-tune the T2I models with paired data. 
However, this training-based method incurs substantial computational cost and suffers from poor generalization.
In contrast, training‑free approaches~\cite{cheng2024scene,zhang2024sketchc,mo2024freecontrol} avoid fine‑tuning by guiding the diffusion denoising process with sketches. 
However, these methods still \emph{jointly} create multiple concepts, causing interference between their differing levels of randomness and abstraction. More critically, both training‑based and training‑free approaches fail to address the \emph{subjective biases} inherent in user‑provided sketches and are particularly vulnerable to highly abstract inputs, often leading to appearance distortions and semantic inconsistencies in the generated outputs. Furthermore, unlike creative generation methods, our focus is on reconstructing real‑world scenes.

\begin{figure*}[!t]
\centering
\includegraphics[width=1.0\textwidth]{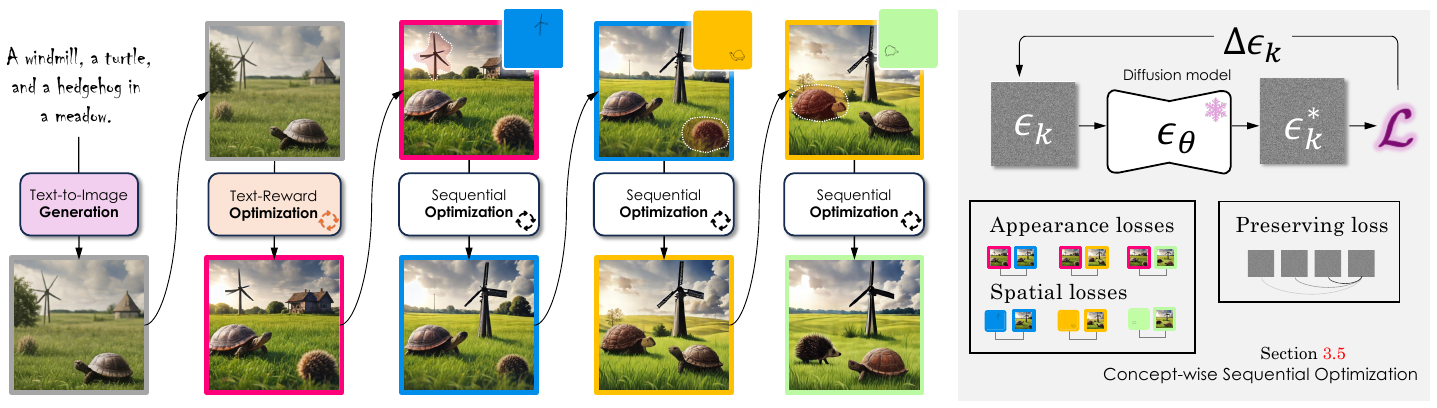}
\caption{
\textbf{Sequence-aware sketch-guided diffusion.}
(\emph{Left}) User-provided textual description undergoes text-to-image generation and text-reward optimization to obtain the initial latent.
Then, the following generation progresses under three constraints --- spatial layout conforming to sketch topology while appearance details adhere to the initial image from text-reward optimization, with consistency maintained through ordered latent propagation and three loss terms, ensuring coherent integration of emerging and established scene elements. 
(\emph{Right}) The concept-wise sequential optimization. 
}
\label{pipeline}
\end{figure*}

\section{Sequence-Aware Sketch-Guided Diffusion}
\label{sec:Method}

\subsection{Overview}

Our goal is to render photorealistic images from subjective readouts, consisting of a textual description $p$ and a sequence $S=\{s_i\}_{i=1}^N$ of $N$ freehand sketches, where each sketch represents a distinct concept. As illustrated in Figure~\ref{pipeline}, we design a Sequence‑Aware Sketch‑Guided Diffusion framework to accomplish this challenging task. Our method processes the subjective readouts individually and sequentially, following the order in which users describe them. It employs a sequence‑aware optimization strategy to align a pre‑trained generative diffusion model, $ \mathcal{G}_:= \epsilon_\theta(\epsilon_\theta(\ldots\epsilon_\theta(\epsilon,p)))$, and $\epsilon_\theta$ denotes the pre-trained U-Net, with these readouts, naturally mirroring the human cognitive process of step‑by‑step scene construction. The process begins with a text‑reward optimization, which establishes an initial latent representation and appearance prior for subsequent concept‑wise optimizations. The appearance prior is extracted via style encoding, while sketch encoding, implemented using diffusion inversion, provides spatial guidance for individual sketches. In addition, a preserving loss is introduced to maintain previously optimized concepts throughout the sequential process. These components collectively enable our framework to decode subjective readouts into coherent, photorealistic scenes.

\subsection{Text-Reward Optimization} 
\label{sec:TRO}

We randomly sample a latent noise $\epsilon$ and feed it, together with the textual description $p$, into a pre‑trained T2I diffusion model, which renders an image through the full reverse diffusion process, $\mathcal{G}_\theta$. To ensure that the generated image contains the concepts in the given textual information, without concept omission or significant misalignment, we formulate a text‑reward optimization procedure as existing practice~\cite{eyring2025reno,lee2024parrot} that iteratively refines the latent noise to maximize a text‑image alignment objective:
\begin{equation}\label{eq:init}
\epsilon^{0\star} = \underset{\epsilon}{\arg \max }\ \mathcal{C}_F\left(\mathcal{G}_\theta(\epsilon, p), p\right),
\end{equation}
where $\mathcal{C}_F$ represents a differentiable reward model that measures alignment between the generated image and the text prompt $p$.
The optimization process iteratively refines $\epsilon$ through gradient ascent:
\begin{equation}
\epsilon_{t+1} = \epsilon_t + \eta \nabla_{\epsilon} \mathcal{C}_F(\mathcal{G}_\theta(\epsilon_t, p)),
\label{noiseoptimization}
\end{equation}
where $\eta$ is the learning rate.

\subsection{Sketch Encoding}
\label{sec:LO}

Before proceeding to the sketch‑guided optimization, we first encode the sketch to extract high‑level semantic guidance, rather than relying on low‑level line details. 
Given a list of input sketches $\{s_i\}_{i=1}^N$, each sketch $s_i$ is encoded into a latent representation $z_i \in \mathbb{R}^{h \times w \times d}$ using the variational autoencoder (VAE) encoder $E_\phi$ of the diffusion model:
$z_i = E_\phi(s_i)$.
Then, we can leverage arbitrary diffusion inversion algorithms, such as DDIM inversion~\cite{mokady2023null} or SDEdit~\cite{meng2021sdedit}, to map the sketch into latent noise. 
To address the significant domain gap between abstract planar sketches and realistic images, we introduce an optimization procedure that iteratively refines the latent representation $z_i$ by aligning it with the rich prior knowledge embedded in the diffusion model. 

At each optimization step $k$, we perturb the latent sketch representation with a Gaussian noise $\epsilon_k \sim \mathcal{N}(0, \mathbf{I})$ to obtain a noisy latent representation
\begin{equation}
\label{addnoise}
z_t^{(k)} = \alpha_{t_k} z_i^{(k)} + \sigma_{t_k} \epsilon_k,
\end{equation}
where $\alpha_{t_k}$ and $\sigma_{t_k}$ are time-dependent scaling factors following the noise schedule of the diffusion process. The noisy latent $z_t^{(k)}$ is then fed into the pre-trained U-Net $\epsilon_\theta$, which predicts the noise component $\hat{\epsilon}_\theta(z_t^{(k)}, t_k)$. The noise prediction error $\Delta \epsilon_k$ is computed as:
\begin{equation}
\Delta \epsilon_k = \hat{\epsilon}_\theta(z_t^{(k)}, t_k) - \epsilon_k.
\end{equation}
This noise gap serves as a gradient signal to iteratively refine the latent sketch features. 
{Specifically, we update the latent representation $z_i^{(k)}$ as}
\begin{equation}
\label{optmization}
z_i^{(k+1)} = z_i^{(k)} - \eta_k \left( \Delta \epsilon_k + \lambda (z_i^{(k)} - z_i^{(0)}) \right),
\end{equation}
where $\lambda$ is a regularization term that anchors the optimization to the initial latent $z_i^{(0)}$, preventing semantic drift. 

Through this iterative optimization in Eq.~(\ref{addnoise}--\ref{optmization}), the latent $\dot{z}_i$is pushed to align with the rich prior knowledge embedded in the diffusion model.\footnote{To avoid ambiguity, we use $\dot{z}_i$ to denote the optimized latent for the $i$‑th sketch concept, distinguishing it from latent $z_i$ would be used in the concept-wise sequential optimization.
} This alignment effectively infuses the sketch with realistic details, textures, and structural coherence, thereby establishing a robust mapping from human hand-drawn sketches to physically grounded visual representations (see Figure~\ref{fig:sketch-opt}).

However, we observed that the optimization method is less effective when the sketch is drawn relatively small. In such cases, the generated result often fails to align with the sketch and instead produces a shape‑independent object in a central position. To address this, we suggest scaling the original sketch using linear interpolation or, alternatively, discarding the optimization for that step.

\begin{figure}
    \centering
    \includegraphics[width=0.88\linewidth]{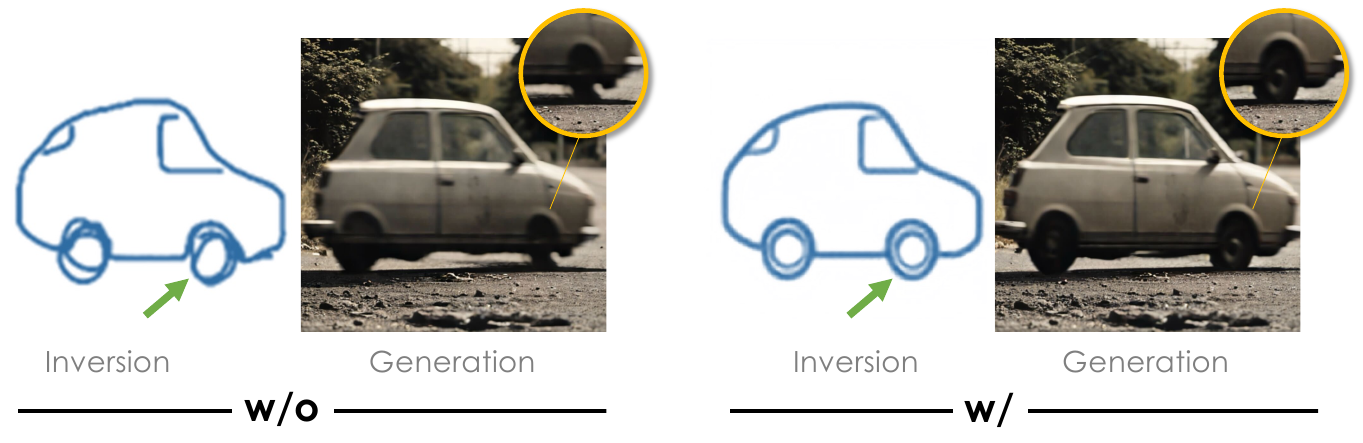}
    \caption{\textbf{The effectiveness of sketch optimization.} Without optimization (\emph{left}), the generated image fails to accurately respect the wheel placement and proportions indicated in the sketch. With our optimization (\emph{right}), the generated result better aligns with the sketch structure, particularly in the wheel positions (green arrows) and overall shape, demonstrating improved spatial coherence.}
    \label{fig:sketch-opt}
\end{figure}

\subsection{Appearance Encoding}
\label{sec:SE}

While the sketch provides spatial information, additional appearance cues are required to ensure visual quality during optimization. 
We first define a operator \(\mathcal{A}(\cdot)\) to obtain the appearance descriptor:
\begin{equation}
\label{appprior}
\mathcal{A}(x) = \left\{
\frac{
\sum_{u,v} \text{sigmoid}\left( \left[ x \right]_{uv} \right) \left[ {f}_x \right]_{uv}
}{
\sum_{u,v} \text{sigmoid} \left( \left[ x \right]_{uv} \right)
} \mid f_x
\right\}\,,
\end{equation}
which is a collection of weighted spatial means of diffusion features \({f}_x\) and \({f}_x\) are from different layer and time steps.
Then, we treat the descriptor extracted from the optimized latent $z^{0\star}$, derived from $\epsilon^{0\star}$ in Eq.~(\ref{eq:init}),
as our appearance prior.
We will compare the appearance descriptor of the current latent and the appearance prior during optimization.

\subsection{Concept-wise Sequential Optimization}
\label{sec:OG}

Next, we leverage this order information of subjective readouts to progressively construct the scene while maintaining coherence between newly introduced and previously established elements.

\heading{{Spatial Attention-guided Control}} 
We leverage the spatial attention mechanisms within the diffusion model's U-Net architecture. For each sketch concept $s_i$, we extract cross-attention maps that highlight regions in the latent space corresponding to specific semantic concepts:
\begin{equation}
\label{eq:kq}
\mathcal{M}(x) = \text{softmax}\left(\frac{Q(x)K(x)^T}{\sqrt{d}}\right) > \tau,
\end{equation}
where $Q(x)$ and $K(x)$ are query and key matrices, $d$ is the dimensionality of the key vectors, and $\tau$ is a threshold parameter. The attention maps $\mathcal{M}(\dot{z}_i)$ extracted from $\dot{z}_i$, the latent of the sketch $s_i$  (Section~\ref{sec:LO}), serve as spatial indicators to guide the concept's layout. The control is given through the spatial loss, defined as:
\begin{equation}
\mathcal{L}_{\text{spatial}}(z_i, \dot{z}_i) \!=\! \frac{\sum_{u,v} [\mathcal{M}(\dot{z}_i)]_{uv} \left\| [\dot{z}_i]_{uv} \!-\! [z_i]_{uv} \right\|_2^2}
     {\sum_{u,v} [\mathcal{M}(\dot{z}_i)]_{uv}},
\end{equation}
where \(u,v\) are spatial positions indices.

\heading{Appearance Control} We define the appearance loss as:
\begin{equation}
    \mathcal{L}_{\text{app}}(z_i, z^{0\star}) = \left\| \mathcal{A}({z_i}) - \mathcal{A}({z^{0\star}}) \right\|_2^2,
\end{equation}
where $\mathcal{A}$ is the operator of exacting appearance descriptors.

\begin{figure*}[t]
  \centering
   \includegraphics[width=1.0\textwidth]{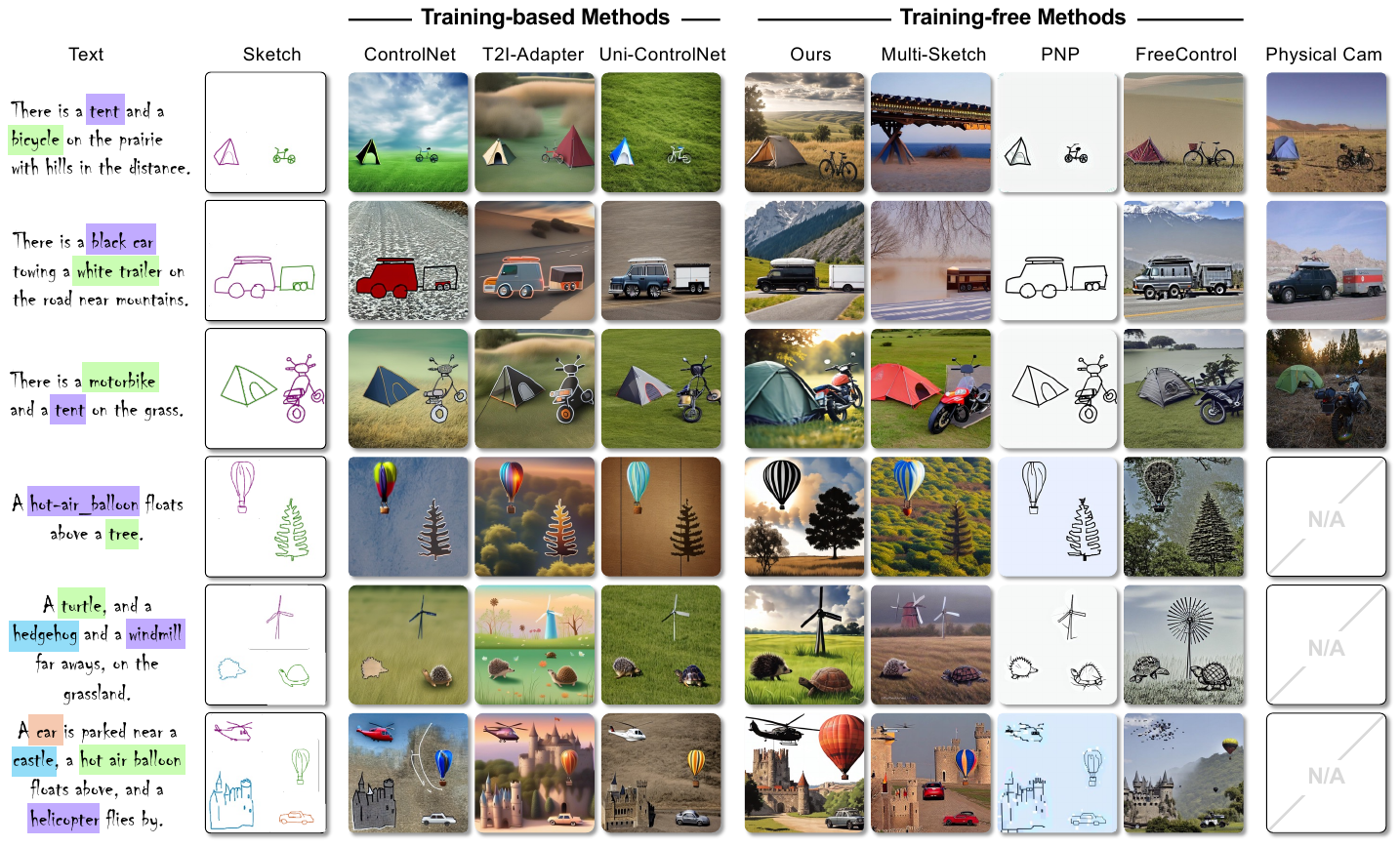}
   \caption{\textbf{Qualitative comparison} of existing sketch-to-image generation methods, and our method on FMC and CMC datasets. 
   Existing training‑based methods tend to overfit to sketch lines, often failing to reconstruct scenes realistically, while training‑free methods frequently misinterpret or overlook key concepts.
   In comparison, our approach more faithfully represents real-world physical scenes. The reading and processing order of the sketches is indicated by colors as follows: \protect\colcirc[1.4ex]{sketch1}{sketch1} \protect\colcirc[1.4ex]{sketch2}{sketch2} \protect\colcirc[1.4ex]{sketch3}{sketch3} \protect\colcirc[1.4ex]{sketch4}{sketch4}.
   }
   \label{fig:comparison}
\end{figure*}

\heading{{Sequential Concept Integration}} To maintain coherence across the progressive addition of concepts, we implement a sequential integration process that updates the latent noise representation while preserving previously incorporated elements. For each sketch $s_i$ in the sequence, we compute:\footnote{In fact, an inner‑loop optimization is omitted here for simplicity.}
\begin{equation}
z_{i} = z_{i-1} + \Delta z_i,
\end{equation}
where $z_{i-1}$ is the optimized latent from the previous concept, and $\Delta z_i$ represents the incremental update required to incorporate the current sketch concept $s_i$.
This update is derived from our loss function: 
\begin{equation}
\label{energy}
\mathcal{L}_i = \alpha_i \mathcal{L}_{\text{new}}(z_i,  \dot{z}_i, z^{0\star}) + \beta_i \mathcal{L}_{\text{preserve}}(z_i, \dot{z}_{0,\ldots, i-1}),
\end{equation}
where $\alpha_i$ and $\beta_i$ are balancing coefficients. $\mathcal{L}_{\text{new}}$ ensures proper integration of the current $z_i$ and is defined as:
\begin{equation}
\mathcal{L}_{\text{new}}(z_i, \dot{z}_i, z^{0\star}) =
\mathcal{L}_{\text{spatial}}(z_i, \dot{z}_i)\ + \mathcal{L}_{\text{app}}(z_i, z^{0\star}),
\end{equation}
while $\mathcal{L}_{\text{preserve}}$ maintains the integrity of previously optimized concepts, given by:
\begin{equation}
\label{prespa}
    \mathcal{L}_{\text{preserve}}(z_i, \dot{z}_{0,\ldots, i-1}) = \sum_{j=1}^{i-1} \gamma^{i-j} \mathcal{L}_{\text{spatial}}(z_i, \dot{z}_j)\,,
\end{equation}
where $j$ means preserving established concepts, and $\gamma^{i-j}$ is a decay factor that weights the importance of earlier concepts based on their step gap from the current step.

\section{Experiments and Results}
\label{sec:Results}

\subsection{Experiment Setup}
\heading{Datasets}
To rigorously evaluate our method, we have undertaken extensive efforts to develop two specialized datasets: the \underline{C}omposed \underline{M}ulti-\underline{C}oncept (CMC) dataset and the \underline{F}reehand \underline{M}ulti-\underline{C}oncept (FMC) dataset. 
Both datasets are designed to comprehensively assess the capabilities of our framework in handling multi-concept scene understanding and generation. 
\textbf{CMC} dataset comprises 142 scene sketches, each containing 2-4 distinct concepts sampled from the 150 semantic classes in the Sketchy dataset~\cite{sangkloy2016sketchy}. 
Each is accompanied by annotated textual prompts with the help of multimodal large language models.
The CMC dataset does not provide real-world images captured by physical cameras for reference.
\textbf{FMC} dataset is constructed from freehand sketches drawn by volunteers, based on real-world images sourced from publicly available outdoor scene datasets. It comprises 42 hand-drawn sketches, each containing 2-3 distinct concepts. Additionally, each sketch is paired with a corresponding real-world image as a reference for validation and evaluation purposes.

\begin{figure*}[t]
  \centering
   \includegraphics[width=0.9\linewidth]{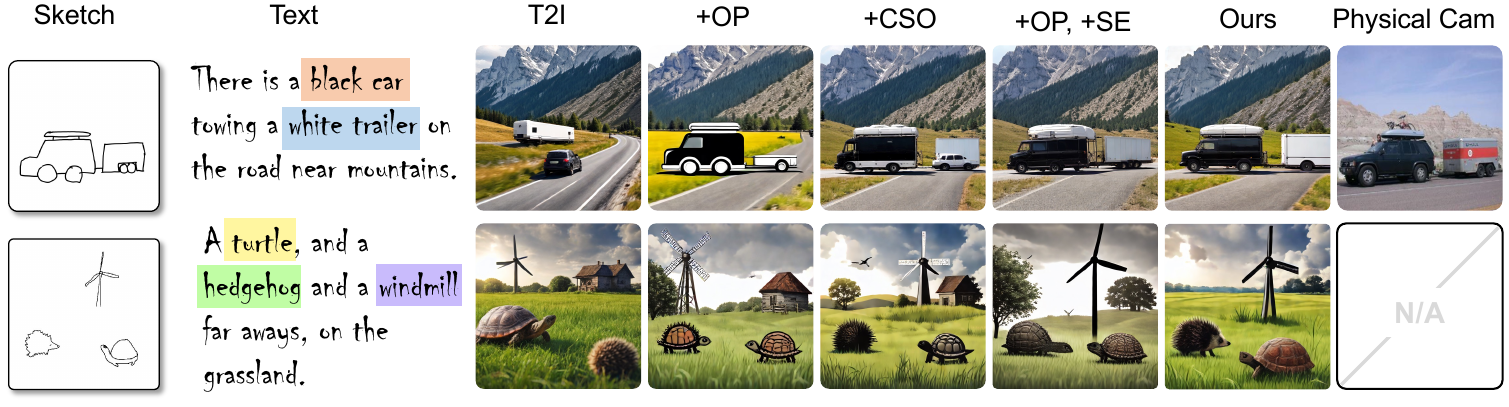}
   \caption{\textbf{Ablation study on FMC and CMC datasets.} Column headers are defined as follows: T2I denotes the baseline text-to-image diffusion model; OP (Overall Perception) indicates the single-step guidance approach utilizing complete sketch as a holistic conditioning signal; CSO refers to our proposed Concept-wise Sequential Optimization (Section~\ref{sec:OG}); SE represents the Sketch Encoding module (Section~\ref{sec:LO}) that enhances structural plausibility by injecting diffusion priors into imperfect sketches. The qualitative comparison demonstrates that CSO achieves finer layout control than OP. While SE effectively bridges the domain gap between freehand sketches and physical reality through latent space regularization, ultimately elevating generation quality. 
} 
   \label{fig:ablation} 
\end{figure*}

\heading{Evaluation Metrics}
To conduct a comprehensive quantitative evaluation of the quality of generated images, four widely-used evaluation metrics were employed, including Fréchet Inception Distance (FID)~\cite{heusel2017gans}, CLIP-T Score (CLIP-T)~\cite{radford2021learning}, Human Preference Score (HPS)~\cite{wu2023human}, and CLIP-I distance (CLIP-I)~\cite{radford2021learning}. 
\textbf{FID} measures the feature distribution similarity between generated and real images. Lower FID values indicate better distribution matching ($\downarrow$). 
\textbf{CLIP-T} quantifies the image-text alignment degree within the CLIP-T embedding space. A higher CLIP-T score reflects greater semantic alignment between the generated image and text prompt ($\uparrow$). \textbf{HPS} measures the subjective quality of the generated images based on human evaluations. Higher HPS values suggest that the generated images are more favorable or preferred by human evaluators~($\uparrow$). 
{
\textbf{CLIP-I} measures the alignment between two images. A higher value suggests that the two images are more consistent in terms of their semantic content ($\uparrow$).
}

\heading{Implementation Details}
In the implementation of our proposed method, we set the learning rate $\eta$ in Eq.~(\ref{noiseoptimization}) and the threshold parameter $\tau$ in Eq.~(\ref{eq:kq}) is 5.0 and 0.3, respectively. Meanwhile, hyperparameters $\alpha_i$ and $\beta_i$ in Eq.~(\ref{energy}) are set as 0.8 and 0.2, respectively. $\gamma$ in Eq.~(\ref{prespa}) is set as 0.9. 
Our method is performed on one NVIDIA GeForce RTX 4090, and input sketch images are 512$\times$512 in resolution.

\begin{table}[t]
  \centering
    \resizebox{\linewidth}{!}
    {
    \begin{tabular}{rccccccc}
    \toprule
    \multirow{2}{*}{\diagbox{Method}{Metric}} & \multicolumn{3}{c}{CMC} & \multicolumn{4}{c}{FMC} \\ 
    \cmidrule(r){2-4} \cmidrule(r){5-8} 
                            & FID $\downarrow$     & CLIP-T $\uparrow$    & HPS $\uparrow$    & FID $\downarrow$  & CLIP-T $\uparrow$  & HPS $\uparrow$  & CLIP-I $\uparrow$  \\ \hline
    \rowcolor{light-gray} \multicolumn{8}{c}{\textcolor{white}{\emph{Training-based methods}}} \\
    ControlNet~\cite{zhang2023adding}              & 15.28 & 61.71 & 0.794 & 14.45 & 66.80 & 0.802 & 0.622 \\
    T2I-Adapter~\cite{mou2024t2i}                   & 18.13 & 61.61 & 0.745 & 15.37 & 61.61 & 0.745 & 0.689 \\ 
    Uni-ControlNet~\cite{zhao2023uni}              & 19.60 & 62.58 & 0.761 & 18.53 & 62.58 & 0.761 & 0.712 \\
    \rowcolor{light-gray} \multicolumn{8}{c}{\textcolor{white}{\emph{Training-free methods}}} \\ 
    Multi-Sketch~\cite{cheng2024scene}             & 13.05 & 64.11 & 0.740 & 11.59 & 72.02 & 0.775 & 0.690 \\
    PNP~\cite{tumanyan2023plug}                    & 17.01 & 63.27 & 0.840 & 16.26 & 71.24 & 0.852 & 0.593 \\
    FreeControl~\cite{mo2024freecontrol}           & 14.52 & 64.74 & 0.757 & 11.17 & 66.67 & 0.759 & 0.775 \\
    Ours                                           & \textbf{11.56} & \textbf{66.64} & \textbf{0.850} & \textbf{10.18} & \textbf{74.25} & \textbf{0.862} & \textbf{0.797} \\
    \bottomrule
    \end{tabular}
    }
    \caption{\textbf{Quantitative results on CMC and FMC datasets.} Our method consistently surpasses all training-free methods in distribution matching, image-text alignment, subjective quality, and appearance details as measured by FID, CLIP-T, HPS, and CLIP-I. Our method achieves superior image-text alignment compared to training-based methods (`$\downarrow$' indicates lower values are better, while `$\uparrow$' is opposite). }
  \label{tab:quantitative}
\end{table}

\subsection{Comparisons}
We conducted a comprehensive quantitative and qualitative evaluation of the proposed method and competing methods on the CMC and FMC datasets. 
Our competing methods include the training‑based approaches ControlNet~\cite{zhang2023adding}, T2I‑Adapter~\cite{mou2024t2i}, and Uni‑ControlNet~\cite{zhao2023uni}, as well as the training‑free approaches Multi‑Sketch~\cite{cheng2024scene}, PNP~\cite{tumanyan2023plug}, and FreeControl~\cite{mo2024freecontrol}.
The quantitative results are presented in Table~\ref{tab:quantitative} and qualitative results are shown in Figure~\ref{fig:comparison}.

We observe that training‑based methods~\cite{zhang2023adding,mou2024t2i,zhao2023uni} are highly susceptible to subjective biases in sketches and often fail to adhere to text prompts, producing distorted and unrealistic images with lower CLIP-T scores. Training‑free methods~\cite{cheng2024scene,hertzprompt,mo2024freecontrol} suffer from concept omissions, conspicuous artifacts, and inconsistencies both among concepts and between concepts and their backgrounds, resulting in poor image realism. These issues stem from mutual interference during the ``snapshot'' generation of multiple concepts.
In contrast, our proposed method faithfully reconstructs real‑world scenes and achieves the best performance across all evaluation metrics on both the CMC and FMC datasets, demonstrating its effectiveness.

\begin{figure}[tbhp]
  \centering
  \includegraphics[width=1.0\linewidth]{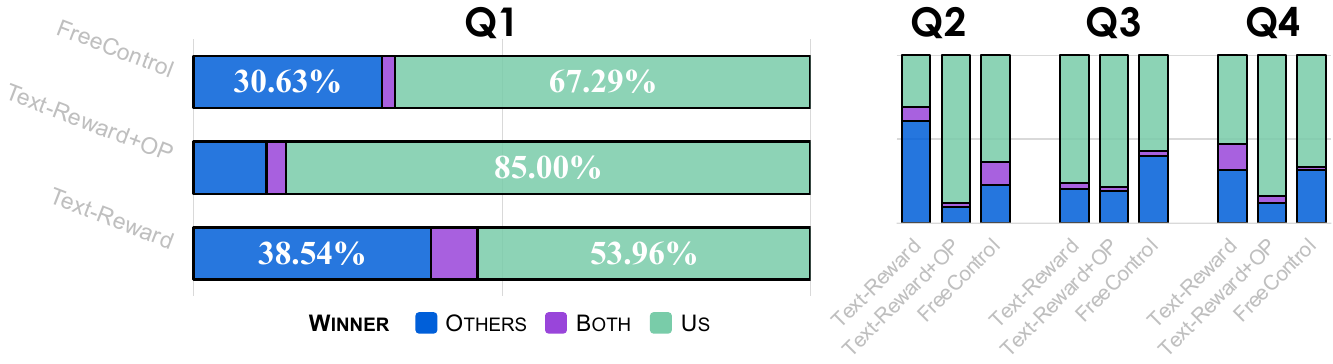}
   \caption{\textbf{User preference experiment.} We asked participants to answer four questions based on the reference images, subjective readouts, and generated images, which included one produced by our method and one by a competing method. Results show that users significantly preferred our outputs over those of competing methods in most cases.
   }
   \label{fig:UPE_overall}
\end{figure}

\subsection{Ablation Study}
To validate the effectiveness of each module in our method, we conducted ablation experiments on the FMC and CMC datasets, with the visualization results shown in Figure~\ref{fig:ablation}. 

It is observed that traditional T2I diffusion models (T2I) generate images that are close to text prompts. While incorporating sketches as guidance further constrains the consistency of concept positioning, deviations in appearance and spatial alignment persist (T2I, OP).
The sequential concept understanding facilitated by Concept-wise Sequential Optimization (CSO) in Section~\ref{sec:OG} enhances the alignment of concept positioning and appearance with input sketches (CSO). Simultaneously, the Sketch Encoding (SE) module in Section~\ref{sec:LO} aligns content latent variables with real-world objects, mitigating geometric misalignment (T2I, OP, SE). When integrating all modules, our method faithfully reconstructs real-world scenes and achieves optimal results (Ours), demonstrating the effectiveness of each module and underscoring the indispensable nature of their design.

\begin{figure}[t]
    \centering
    \includegraphics[width=1\columnwidth]{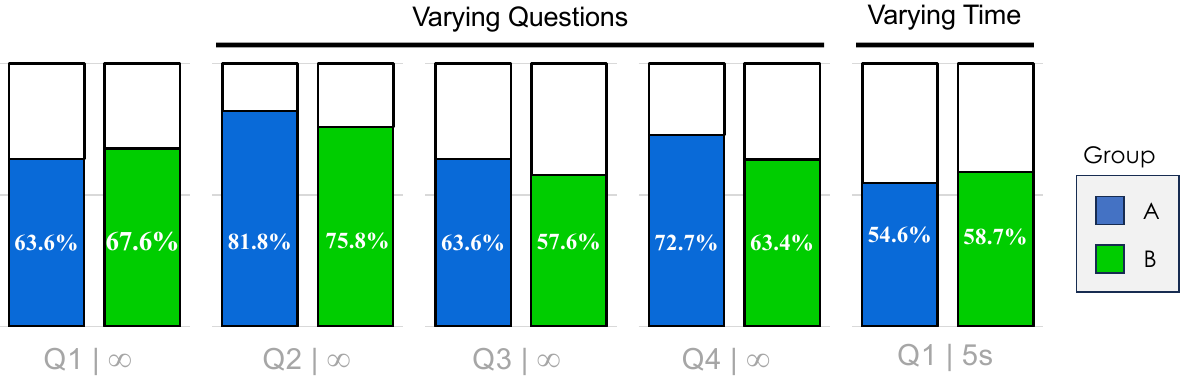}  %
    \caption{\textbf{Control experiments on user preferences.} We controlled three factors to examine their influence: whether the rater drew the photo, the availability timing of the reference image ($\infty$ \vs 5s), and the types of questions asked. In general, our method is consistently better than the competing method, FreeControl.
    }
    \label{fig:UPE}
\end{figure}

\subsection{User Preference Experiments}

We invited 12 participants (group A) to provide subjective readouts and an additional 28 participants (group B) to evaluate the reconstructed images. Participants in group A were first asked to select a deeply impressive scene from their photo gallery and then provide a textual description along with sketches within two minutes. Our method and competing approaches, including two baselines (text‑reward optimization and text‑reward optimization with sketch‑guided optimization based on all concepts) as well as an optimal method, FreeControl~\cite{mo2024freecontrol}, were used to generate images from these readouts. 
Participants in both groups were then shown the subjective readouts, the reference image, and the generated images for the same scene. They were asked to answer four single‑choice questions in a single‑blind manner:  
Q1: Which image is closer to the original image (both style and content)?  
Q2: Which image has higher visual quality (\eg, more realistic, more natural)? 
Q3: Which image better matches the structure of the sketch?  
Q4: Which image better matches the textual description?  

Figure~\ref{fig:UPE_overall} presents the statistics of the collected results. Participants judged our reconstructions as being closer to the corresponding real‑world scenes compared to competing methods. Similar conclusions were observed for the other three questions. However, there was an exception in the comparison between our method and text‑reward optimization regarding image quality (Q2). We observed that text‑reward optimization without additional control tends to hallucinate details that enhance perceived realism, leading some users to prefer it for Q2. This also explains why our advantage over text‑reward optimization in Q1 was not statistically significant and why some participants expressed hesitation in this case.

We further conducted controlled experiments to examine the effects of three factors: whether the rater provided the subjective readouts, the availability timing of the reference image, and the types of questions asked. Results in Figure~\ref{fig:UPE} show that our method consistently outperformed FreeControl~\cite{mo2024freecontrol}, regardless of these variations. Notably:  
1) Participants who provided subjective readouts showed slightly lower preference for our method in terms of similarity to the reference image compared to those who did not, while showing a higher preference in terms of alignment with the readouts and perceived image quality. We suppose this is mainly because of the subjectivity in reading out.
2) Reducing the time participants could view the reference image decreased the preference for our method; when allowed to compare freely, participants preferred our results more strongly.

\section{Conclusions}

This paper presented {Subjective Camera}, a paradigm for reconstructing real‑world scenes from humans' mental impressions. Our framework relies on the most accessible readouts of mental imagery, \ie, textual descriptions and sketches, and employs a training‑free, sequence‑aware optimization process. This approach eliminates the need for costly fine‑tuning of text‑to‑image diffusion models on large‑scale data while mitigating challenges such as dependence on sketch quality and geometric misalignment.
Extensive experiments on both synthetic and real‑world sketch datasets demonstrate that {Subjective Camera} achieves state‑of‑the‑art performance, outperforming both training‑free and training‑based sketch‑to‑image generation methods across all evaluation metrics. Furthermore, user preference studies confirm that participants consistently favor our method.

In closing, we address a natural question: ``With the rapid advancement of generative AI, will physical cameras ever become unnecessary'' For now, the answer is \emph{no}. First, the reconstructed images still fall short of perfectly matching actual scenes. This limitation largely stems from the fact that current T2I models are trained on highly diverse data, creating gaps between the prior and the specific scene to be reconstructed. One promising direction to mitigate this is to personalize the prior models using user‑specific galleries. Second, we observe that variations in the order and quality of sketches have a significant impact on reconstruction results. To address this, we plan to enhance our framework with computational tools that guide users in providing more precise and suitable subjective readouts. For these reasons, we designate this system as {Subjective Camera 1.0}, leaving these improvements for future iterations.

\heading{Acknowledgements} This work was supported in part by the National Natural Science Foundation of China (62225113, 624B2109, 623B2079) and the Zhongguancun Academy Project (20240308). We thank Po‑Hsun Huang for providing the questionnaire system for our user study, Kaijin Zheng for assistance with data processing, and all participants who took part in the studies.

{
    \small
    \bibliographystyle{ieeenat_fullname}
    \bibliography{main}
}

\end{document}